\documentclass[11pt]{article}

\usepackage[final]{acl}

\usepackage{times}
\usepackage{latexsym}

\usepackage[T1]{fontenc}

\usepackage[utf8]{inputenc}

\usepackage{microtype}

\usepackage{inconsolata}

\usepackage{graphicx}

\usepackage{tcolorbox}

\usepackage{makecell}

\usepackage{float}

\title{When Rating Scales Fall Short: LLM-Assisted Discovery of ADHD Signals in Turkish Teacher Narratives}

\author{
\textbf{Baris Karacan\textsuperscript{1}\thanks{Corresponding author}},
\textbf{Irem Aktar Songur\textsuperscript{2}},
\textbf{Ahmet Ozaslan\textsuperscript{2}},
\textbf{Elvan Iseri\textsuperscript{2}}
\\
\textsuperscript{1}Department of Computer Science, University of Illinois Chicago
\\
\textsuperscript{2}Department of Child and Adolescent Psychiatry, Gazi University
\\
\texttt{bkarac3@uic.edu},
\texttt{\{driremaktar,ozaslanahmet,iserielvan\}@gmail.com}
\\
}

\begin{document}
\maketitle
\begin{abstract}
Attention Deficit Hyperactivity Disorder (ADHD) is one of the most common neurodevelopmental disorders in childhood, and its diagnosis relies on assessments combining clinician judgment with standardized rating scales and reports from parents and teachers. While structured instruments such as the Conners' Teacher Rating Scale–Revised Short Form (CTRS-R:S) quantify ADHD-related behaviors, teachers also provide open-ended narratives that may contain complementary signals not captured by structured assessments. However, it remains unclear to what extent teacher narratives encode signals overlooked by rating scales. In this study, we analyze de-identified Turkish teacher evaluation forms collected during clinical ADHD assessments, including both CTRS-R:S scores and open-ended teacher narratives. We compare predictive signals from structured scores and narrative text and identify cases where structured assessments fail to clearly distinguish ADHD from non-ADHD students while narrative-based models capture distinct behavioral patterns. Notably, these cases show minimal overlap with those missed by the narrative model, suggesting that structured and narrative information encode complementary signals. To interpret these differences, we apply a large language model (LLM)-assisted theme discovery pipeline that reveals distinct attention, behavioral, and family-related patterns, highlighting the potential of natural language processing (NLP) to uncover clinically relevant signals from teacher narratives and to complement traditional ADHD screening tools.
\end{abstract}

\section{Introduction}
Attention Deficit Hyperactivity Disorder (ADHD) is one of the most common neurodevelopmental disorders in childhood, with global prevalence estimates exceeding 5\%~\citep{drechsler2020adhd} and epidemiological studies in Turkey reporting rates up to 13\%~\citep{ercan2013prevalence}. ADHD is characterized by persistent patterns of inattention, hyperactivity, and impulsivity that are inconsistent with developmental expectations and can impair functioning across multiple settings, including school and social environments~\citep{american2022diagnostic, skounti2007variations, shaw2012systematic}. Although emerging diagnostic tools and biomarkers, such as quantitative electroencephalography (qEEG), have shown promise~\citep{byeon2020novel, kopanska2025aberrant, duric2025quantitative}, they cannot be used as standalone diagnostic tools. Thus ADHD diagnosis continues to rely on clinical evaluation supported by information from multiple sources, including standardized behavioral rating scales and reports from parents, teachers, and the child~\citep{hall2016clinical, peterson2024tools}. Among these informants, teachers play an important role because they observe children's behavior in structured learning environments where attention, impulse control, and peer interactions are continuously evaluated~\citep{ford2008five, malpica2014manual}.  


In clinical practice, teacher reports are often collected using both structured rating scales and open-ended evaluation forms. One commonly used instrument is the Conners' Teacher Rating Scale–Revised Short Form (CTRS-R:S), a 28-item questionnaire rated on a four-point Likert scale that measures symptoms such as cognitive problems–inattention, hyperactivity, and oppositional behaviors~\citep{conners1998revision}. In this study, we use the Turkish adaptation of the scale, which has undergone validity and reliability evaluation~\citep{kaner2013conners}. In addition to such structured assessments, teachers often provide qualitative descriptions through open-ended evaluation forms, offering contextual observations about students' behavior that may not be fully captured by structured rating scales. These forms typically gather information across several areas of a student's functioning, including school performance, prominent personality characteristics and perceived mental health concerns, relationships with teachers and peers, and teachers’ observations about the family context. While structured rating scales such as the CTRS-R:S are widely used to quantify ADHD-related behaviors, the qualitative descriptions provided by teachers remain less systematically analyzed. These teacher narratives often contain contextual information about how behavioral difficulties manifest in classroom environments. However, such descriptions are typically reviewed qualitatively during clinical assessment and are rarely examined using structured or computational methods. As a result, potentially informative behavioral signals contained in teacher narratives remain underexplored. This gap suggests an opportunity for NLP methods to analyze teacher narratives at scale and uncover behavioral patterns that may not be explicitly captured by structured rating scales.

In this work, we analyze teacher evaluation forms collected during clinical ADHD assessments that include structured rating scales and open-ended teacher narratives. We compare predictive signals from structured CTRS-R:S scores and narrative text to examine how each modality distinguishes ADHD from non-ADHD students and whether the two sources capture complementary behavioral patterns. We contribute:

\begin{enumerate}
    \item \textbf{Joint analysis of structured and narrative teacher reports. } We conduct a comparative analysis of structured CTRS-R:S scores and narrative teacher descriptions to examine how each modality identifies students with ADHD.
    \item \textbf{Complementary predictive signals across modalities.} Through error analysis, we show that models trained on structured rating scores and narrative teacher reports miss partially distinct cases, revealing minimal overlap between false negatives across the two modalities, and suggesting that the two sources capture complementary information.
    \item \textbf{LLM-assisted theme discovery for interpretability.} To better understand the differences between modalities, we introduce an LLM-based pipeline that extracts behavioral themes from teacher narratives and organizes them into expert-validated higher-level categories.
    \item \textbf{A dataset of Turkish teacher evaluation forms. } We release a de-identified dataset of teacher reports from 199 students (166 ADHD, 33 control) collected during clinical ADHD assessments, providing a resource for computational analysis of behavioral narratives in a less-studied language.
\end{enumerate}


\section{Related Work}

Natural language processing (NLP) has increasingly been applied to the study of mental health and psychiatric conditions using textual and conversational data. Prior work has explored the detection of mental health signals from a variety of sources, including social media posts, electronic health records (EHR), and transcribed patient–doctor interactions. For example, \citet{coppersmith2014quantifying} analyzed Twitter data to detect linguistic signals associated with mental health conditions such as bipolar disorder, major depressive disorder, and post-traumatic stress disorder. In clinical settings, \citet{rumshisky2016predicting} applied topic modeling to EHR data to identify patients at high risk of psychiatric readmission. Linguistic markers extracted from speech have also been shown to predict psychiatric outcomes such as psychosis onset~\citep{bedi2015automated}. More recently, \citet{karacan2024towards} examined clinically elicited speech to identify linguistic markers associated with psychiatric conditions including bipolar disorder and schizophrenia. 

Together, these studies demonstrate the potential of language-based analysis to uncover behavioral and psychological patterns relevant to mental health. However, most prior NLP research in mental health focuses on patient-generated language (e.g., social media posts or self-reports) or clinician-authored clinical documentation, often in adult populations. In contrast, behavioral observations written by teachers during child and adolescent assessments—an important component of ADHD evaluation—remain largely unexplored in computational studies. This gap motivates the present study, which applies NLP methods to teacher narratives collected during clinical ADHD assessments and compares signals from narrative descriptions with structured behavioral rating scales.

\section{Dataset}
The dataset used in this study was collected from the Child and Adolescent Psychiatry outpatient clinic at Gazi University Faculty of Medicine between 2019 and 2024. The study protocol was approved by the institution’s ethics review board. All records were retrospectively reviewed and de-identified prior to analysis. The ADHD group consists of children and adolescents diagnosed with attention-deficit/hyperactivity disorder according to ICD-10 diagnostic codes F90.0–F90.9. For patients with repeated diagnoses during follow-up, only the initial diagnosis was considered. Teacher evaluation forms and Conners’ Teacher Rating Scale–Revised Short Form (CTRS-R:S) assessments collected at that time were included in the study. Patients with other psychiatric diagnoses or incomplete records were excluded. 

The control group consists of children and adolescents without active psychopathology who were evaluated in the same clinic with ICD-10 diagnostic codes Z00.4 and Z71.0. Only cases with both teacher evaluation forms and CTRS-R:S scores were included. Individuals with prior psychiatric diagnoses or chronic physical illnesses were excluded. Controls were clinic-referred rather than community-recruited to ensure that the absence of active psychopathology was clinically assessed within the same care setting as the ADHD group. This design also allowed the absence of active psychopathology to be clinically assessed, which would be less feasible in an unverified community control group. The final dataset contains 199 students aged 6-14 years (166 ADHD, 33 control)\footnote{The de-identified dataset is available at: \\ \url{https://github.com/barikosan/teacher-narrative-adhd-dataset/}}. For each student, the dataset includes both structured behavioral ratings from the CTRS-R:S and teacher-written narrative reports.

\subsection{Conners’ Teacher Rating Scale–Revised Short Form (CTRS-R:S)}
The Conners’ Teacher Rating Scale–Revised Short Form (CTRS-R:S) is a widely used teacher-reported instrument for assessing ADHD-related behaviors in classroom settings. The scale contains 28 behavioral items, each rated on a four-point Likert scale, where the options \textit{“never,”} \textit{“rarely,”} \textit{“frequently,”} and \textit{“always”} are scored as 0, 1, 2, and 3, respectively. These items describe observable behaviors related to attention, impulsivity, classroom conduct, and academic functioning.

The scale is organized into three primary subscales—Oppositional Defiant Behavior, Cognitive Problems–Inattention, and Hyperactivity—as well as an auxiliary ADHD Index used for screening purposes. Example items include behaviors such as \textit{difficulty sustaining attention, leaving one’s seat in situations where remaining seated is expected,} and \textit{interrupting or intruding on others}. In this study, we use the validated Turkish adaptation of the CTRS-R:S described by \citet{kaner2013conners}.

\subsection{Open-Ended Teacher Evaluation Form}
In addition to structured ratings, teachers provide qualitative descriptions through an open-ended evaluation form. The form collects narrative observations across four areas:
\begin{itemize}
    \item \textbf{School performance}, including academic strengths, difficulties, success relative to ability, and areas of interest.
    \item \textbf{Personality characteristics and perceived mental health concerns.}
    \item \textbf{Relationships with teachers and peers.}
    \item \textbf{Teacher observations regarding the family context.}
\end{itemize}

These narratives provide contextual behavioral descriptions that complement structured ADHD rating scales and constitute the primary textual data analyzed in this study. An example excerpt from a teacher narrative related to the \textit{School Performance} section is shown in Figure~\ref{fig:quote}.


\begin{figure}[t]
\centering
\footnotesize
\begin{minipage}{0.97\linewidth}
\begin{tabular}{|p{0.97\linewidth}|}
\hline
\textbf{OKUL BAŞARISI (Original Turkish):} \\
"Derslere aktif katılım göstermez, sadece sorulduğunda cevap verir. Genellikle derslerde eşyalarıyla ilgilenir. Özel bir ilgi alanı veya yeteneği gözlenmemiştir. Akademik başarısı orta seviye olarak değerlendirilebilir." \\
\hline
\hline
\textbf{SCHOOL PERFORMANCE (English Translation):} \\
"Does not actively participate in class and typically responds only when called upon. During lessons, the student often focuses on personal belongings rather than classroom activities. No particular academic interests or special skills have been observed. Overall academic performance is average." \\
\hline
\end{tabular}
\caption{Example teacher narrative from the school performance section for a student diagnosed with ADHD. The original Turkish text is shown alongside its English translation for reference; all analyses were conducted on the original Turkish text.}
\label{fig:quote}
\end{minipage}
\end{figure}

\section{Methodology}
\label{sec:methods}

Our methodology consists of two main stages. In the first stage, we construct two classification pipelines to examine predictive signals derived from different modalities of teacher reports. The first pipeline uses structured behavioral ratings from the Conners’ Teacher Rating Scale–Revised Short Form (CTRS-R:S) to train a classifier based on standardized ADHD assessment scores. The second pipeline analyzes teacher-written narratives using NLP techniques to represent the text and train a text-based classifier.

In the second stage, we conduct an error-focused qualitative analysis to better understand behavioral patterns captured in teacher narratives. Specifically, we examine cases that are misclassified by the structured model and analyze their corresponding teacher narratives using an LLM-assisted theme identification procedure. The extracted themes are then used to characterize behavioral patterns present in these cases and to compare them with those observed in the control group. This analysis provides qualitative insights into behavioral characteristics that may not be fully captured by structured rating scales.

\subsection{Structured Classification Pipeline}
To model behavioral signals captured by standardized teacher assessments, we constructed a structured classification pipeline using features derived from the Conners’ Teacher Rating Scale–Revised Short Form (CTRS-R:S). The primary feature set consists of the 28 behavioral rating items included in the CTRS-R:S, which reflect observable classroom behaviors related to attention, impulsivity, and oppositional conduct. In addition to these behavioral ratings, we include basic demographic variables—\textit{student age}, \textit{grade level}, and \textit{gender}—as contextual features. These variables provide information about the student’s developmental stage and classroom expectations, which can influence how behavioral symptoms are observed and reported by teachers.

Before model training, numeric variables (age, grade, and CTRS-R:S item scores) are standardized using z-score normalization, while the categorical gender variable is encoded using one-hot encoding. These preprocessing steps are integrated into the training pipeline to prevent data leakage during cross-validation. We then train a logistic regression classifier with class-balanced weighting to account for the class imbalance present in the dataset. Logistic regression provides a simple baseline for evaluating predictive signals in structured behavioral ratings. Given the relatively small and imbalanced dataset, we favor a linear model over more data-intensive approaches (e.g., transformer-based models), which typically require larger labeled datasets to generalize reliably.


Model evaluation is performed using stratified five-fold cross-validation to preserve the class distribution across folds. For each fold, the model generates out-of-fold predictions, ensuring that each instance is evaluated by a model that was not trained on that instance. These predictions are later used to examine error patterns across modalities. In particular, they allow us to identify cases misclassified by the structured model and analyze how these instances differ from those captured by narrative-based signals in subsequent analyses.

\subsection{Narrative Text Classification Pipeline}
\label{sec:narrative}

To model behavioral signals expressed in teacher narratives, we construct a text-based classification pipeline using NLP techniques. Each teacher evaluation form contains free-text descriptions written by teachers, capturing behavioral observations that may not be explicitly represented in standardized rating scales. Prior to feature extraction, narrative texts undergo light preprocessing to normalize the input representation. This includes lowercasing and removal of common Turkish stopwords using a publicly available stopword list~\citep{korkmaz2023turkishstopwords}. These steps reduce noise while preserving the original lexical content of teacher descriptions.

After preprocessing, narratives are represented using a bag-of-words representation with term frequency–inverse document frequency (TF–IDF) weighting~\citep{salton1988term}. We use unigram and bigram features, retaining only terms that appear in at least two documents and excluding terms that appear in more than 95\% of documents to reduce sparsity and remove overly common terms. Using these TF–IDF features, we train a logistic regression classifier to predict ADHD status from teacher narratives, following the same modeling approach used in the structured classification pipeline.

Similarly, model evaluation follows the same stratified five-fold cross-validation procedure described in the structured classification pipeline. This ensures that predictions from both modalities are produced under the same evaluation setup. Out-of-fold predictions generated during cross-validation are later used to analyze differences between structured and narrative signals, particularly in cases where narrative-based predictions correctly identify instances misclassified by the structured model.

\subsection{LLM-Assisted Theme Identification Pipeline}
\label{sec:llm}

To better understand behavioral patterns expressed in teacher narratives, we introduce an LLM-assisted theme identification pipeline that extracts recurring behavioral descriptions from narrative text. While the structured classifier successfully identifies many ADHD cases, some instances remain difficult to detect using standardized rating scales alone. In several of these borderline cases, the narrative-based classifier captures behavioral cues that are not reflected in structured ratings (see Section~\ref{sec:experiments} for detailed analysis). However, the narrative classifier only produces quantitative predictions and does not directly reveal the types of behavioral observations described in teacher reports. 

We interpret these differences between modalities by applying an LLM-based theme extraction procedure to a subset of teacher narratives. Specifically, the analysis focuses on instances that are misclassified by the structured classifier as well as narratives from the control group. The procedure extracts concise behavioral themes from teacher descriptions and organizes them into recurring patterns. By comparing themes observed in these instances with those present in the control group, we aim to uncover behavioral signals that may not be explicitly captured by standardized rating scales.

Although teacher narratives are de-identified, they may still contain sensitive information about students. Using closed-source API-based models could therefore introduce potential privacy and security risks when processing clinical text~\citep{kim2025benchmarking}. To mitigate these concerns, we employ the open-weight \textit{Llama 3.3 70B Instruct} model~\citep{grattafiori2024llama}, which can be deployed locally without transmitting data to external services. In addition to privacy considerations, the model provides strong instruction-following capabilities and multilingual generalization, which is important for analyzing Turkish teacher narratives written in free text. The model also offers a large context window (128K tokens), allowing complete teacher reports to be processed without truncation. Theme extraction was performed using deterministic decoding with a fixed maximum output length of 128 tokens.

Themes are extracted using a structured prompt-based procedure consisting of a system instruction and a user instruction. The prompt instructs the model to extract short behavioral themes grounded in the teacher narrative while avoiding diagnostic labels or speculative interpretations. Each theme must describe an observable behavioral strength or difficulty reported by the teacher. 

The extraction is organized around four domains corresponding to the teacher evaluation form: \textit{school performance}, \textit{personality or mental health characteristics}, \textit{social relationships}, and \textit{family context}. To encourage balanced coverage, the model may extract at most two themes per domain (maximum eight themes per narrative) and returns the results in a structured JSON format including the theme text, its valence (positive or negative), and the domain label. Figure~\ref{fig:prompt} presents an English translation of the prompt used for theme extraction, while the original Turkish prompt template used in the experiments is provided in Appendix~\ref{sec:appendix1}.

\begin{figure}[!t]
\centering
\begin{tcolorbox}
\scriptsize
\textbf{SYSTEM:} You are not a clinical diagnostician. \\
Extract short behavioral themes from teacher narratives that describe observable student behaviors. \\

\textbf{Rules:} \\
- Do not produce diagnostic labels (e.g., ADHD), diagnoses, reports, or medication references. \\
- Each theme must contain 2--6 words. \\
- Each theme must clearly describe either a \textit{strength (positive)} or a \textit{difficulty (negative)}. \\
- Do not produce neutral themes. \\
- Only extract themes explicitly supported by the text. \\
- Return only valid JSON. \\

\textbf{Example themes:} \\
\textit{"strong social adjustment"} \\
\textit{"difficulty with peer relationships"} \\

\textbf{USER:} 
Read the teacher narrative below and extract up to two themes for each domain: \\

1) school performance \\
2) personality / mental characteristics \\
3) social relationships \\
4) family context. \\

\textbf{Output JSON format:}

\begin{quote}
\ttfamily
\{\\
\hspace*{1em}"themes": [\\
\hspace*{2em}\{\\
\hspace*{3em}"theme": "...",\\
\hspace*{3em}"valence": "positive|negative",\\
\hspace*{3em}"domain": "school|personal|social|family"\\
\hspace*{2em}\},\\
\hspace*{2em}...\\
\hspace*{1em}]\\
\}
\end{quote}

\textbf{Additional rules:}
At most two themes per domain, at most eight themes in total, and all themes must be grounded in the narrative text. \\

\textbf{Teacher narrative:} \texttt{<<< \{TEXT\} >>>}
\end{tcolorbox}
\caption{Prompt used for LLM-assisted theme extraction from Turkish teacher narratives (English translation).}
\label{fig:prompt}
\end{figure}

Since the prompt allows the model to extract themes freely within the four domains of the teacher evaluation form, the resulting themes often vary in granularity. For example, themes such as \textit{"çekingen kişilik"} (shy personality) and \textit{"içe kapanıklık"} (social withdrawal) both describe introverted behavior and can be interpreted under a broader category of \textit{"social-emotional difficulty"}. 

To improve interpretability and enable systematic comparison across students, extracted themes are manually mapped to higher-level behavioral categories in consultation with a domain expert in child and adolescent psychiatry. This mapping results in ten fundamental categories that capture common strengths and difficulties described in teacher narratives: \textit{attention difficulty}, \textit{academic difficulty}, \textit{behavioral difficulty}, \textit{social-emotional difficulty}, \textit{academic strength}, \textit{behavioral strength}, \textit{social-emotional strength}, \textit{family support}, \textit{family difficulty}, and \textit{neutral context}\footnote{ A few outputs were descriptive and lacked clear polarity; these were mapped to \textit{neutral context}}.

This categorization produces a structured representation of behavioral signals contained in teacher narratives, enabling systematic comparison between ADHD cases that were misclassified by the structured model and the control group. Importantly, the LLM-assisted procedure is not intended to replace predictive modeling but rather to provide qualitative insights into behavioral descriptions present in teacher narratives. Therefore, we used the LLM strictly as an interpretive tool rather than a diagnostic classifier.

\section{Experiments}
\label{sec:experiments}

\subsection{Experimental Setup}
Both the structured and narrative classification pipelines are evaluated under the same stratified five-fold cross-validation setup described in Section~\ref{sec:methods}. This shared evaluation protocol ensures that the two modalities can be compared under identical training and testing conditions. We report area under the precision–recall curve (PR-AUC), area under the receiver operating characteristic curve (ROC-AUC), balanced accuracy, and recall for the ADHD class. PR-AUC is particularly informative under class imbalance, while ADHD recall directly reflects the model's ability to detect ADHD cases and avoid false negatives. All classification models were implemented using the scikit-learn library~\citep{pedregosa2011scikit}.

In addition to overall predictive performance, we conduct error-focused follow-up analyses to better characterize ADHD cases missed by the structured classifier (i.e., structured false negatives). These analyses compare structured false-negative cases with controls using (i) statistical tests over structured rating scores, (ii) subset classification analyses over teacher narratives, and (iii) association analyses between structured and text-based prediction scores. Together, they assess whether structured false negatives resemble controls under standardized ratings while remaining distinguishable in narrative text, thereby motivating the subsequent LLM-assisted theme analysis.

\subsection{Classification Performance Across Modalities}
We first compare the overall classification performance of the structured and narrative pipelines. Table~\ref{tab:classification} reports the predictive performance of the two pipelines. Both approaches achieve strong discrimination between ADHD and control groups, with high PR-AUC values above 0.98. While the structured pipeline shows higher balanced accuracy, the narrative classifier achieves substantially higher recall for ADHD cases (0.958 vs. 0.856), suggesting that teacher narratives capture behavioral signals that may not be fully reflected in standardized rating scales. 

Consistent with this observation, the structured classifier produces 24 false-negative ADHD cases (14.5\% of ADHD instances), whereas the narrative classifier misses only 7 cases (4.2\%). Notably, the overlap between the two false-negative sets is minimal, with only one shared instance. This pattern suggests that the two modalities capture partially distinct behavioral signals, motivating further analysis of cross-modal error patterns.

\begin{table}[t]
\centering
\small
\setlength{\tabcolsep}{3pt}
\begin{tabular}{lcccc}
\hline
\textbf{Model} & \textbf{PR-AUC} & \textbf{ROC-AUC} & \textbf{Bal. Acc.} & \textbf{ADHD Rec.} \\
\hline
\hline
Structured &
\makecell{0.985\\$\pm$ 0.012} &
\makecell{0.921\\$\pm$ 0.062} &
\makecell{\textbf{0.799}\\$\pm$ 0.114} &
\makecell{0.856\\$\pm$ 0.060} \\
\hline
Narrative &
\makecell{\textbf{0.989}\\$\pm$ 0.006} &
\makecell{\textbf{0.944}\\$\pm$ 0.033} &
\makecell{0.722\\$\pm$ 0.113} &
\makecell{\textbf{0.958}\\$\pm$ 0.045} \\
\hline
\end{tabular}
\caption{Classification performance of logistic regression models trained on structured ratings and teacher narratives under stratified five-fold cross-validation. Values are reported as mean $\pm$ standard deviation. Bal. Acc. = balanced accuracy; ADHD Rec. = ADHD recall.}
\label{tab:classification}
\end{table}

\subsection{Quantitative Error Analysis Across Modalities}
\label{exp:5.3}
We first examine whether ADHD cases missed by the structured classifier resemble control cases under structured behavioral ratings. For each student, we compute a summary structured score defined as the mean of the 28 CTRS-R:S behavioral rating items. This score provides an aggregate measure of the severity of observable classroom behaviors reported by teachers. The average structured score for correctly classified ADHD cases is substantially higher (1.48) than for structured false-negative cases (0.61), while control cases exhibit the lowest scores on average (0.39). Given that individual rating items are scored on a 0-3 scale, this pattern suggests that ADHD cases missed by the structured classifier exhibit milder behavioral signals in standardized rating scales. To statistically evaluate this observation, we compare structured false-negative cases  ($n = 24$) with control cases ($n = 33$) using a two-sided Mann–Whitney U test~\citep{macfarland2016mann}. The difference between the two groups is statistically significant ($p = 0.014$), indicating that structured false negatives still exhibit somewhat elevated behavioral scores compared to controls, although the separation between these groups is relatively limited.

To further quantify the discriminative power of structured scores within this subset, we compute the ROC-AUC using the structured severity score as the predictor while considering only structured false-negative ADHD cases and controls. The resulting ROC-AUC of 0.69 indicates that structured ratings provide only moderate discrimination between these two groups. This finding suggests that ADHD cases missed by the structured classifier exhibit rating patterns that overlap substantially with controls, making them difficult to distinguish using structured scores alone.

We next examine whether behavioral signals expressed in teacher narratives can help distinguish these difficult cases. Using the narrative text representation described in Section~\ref{sec:narrative}, we evaluate the ability of teacher narratives to separate structured false-negative ADHD cases from control students. When restricting the analysis to this subset ($n = 57$), we train a logistic regression classifier using the TF--IDF representation under the same stratified five-fold cross-validation setup. The narrative-based classifier achieves a ROC-AUC of 0.84, substantially higher than the ROC-AUC obtained using structured severity scores alone (0.69). This result indicates that teacher narratives contain behavioral signals that remain informative even when standardized rating scores appear similar to those of control students.


To further examine the relationship between the two modalities, we analyze the association between structured severity scores and narrative-based prediction probabilities within the same subset. The Pearson correlation between these two signals is very weak ($r = 0.07$), suggesting that structured ratings and narrative descriptions capture largely distinct behavioral information.

Taken together, these findings provide quantitative evidence that teacher narratives contain complementary behavioral signals beyond those captured by standardized rating scales. This motivates a closer qualitative examination of behavioral descriptions in teacher narratives, which we explore using the LLM-assisted theme identification pipeline described in Section~\ref{sec:llm}.

\subsection{LLM-Assisted Qualitative Theme Analysis}
To better understand the behavioral signals present in teacher narratives, we apply the LLM-assisted theme extraction pipeline described in Section~\ref{sec:methods}. The analysis focuses on the subset examined in Section~\ref{exp:5.3}, consisting of structured false-negative ADHD cases ($n = 24$) and control students ($n = 33$). For each teacher narrative, the LLM extracts concise behavioral themes that describe observable student behaviors and contextual information reported by teachers. The extracted themes are mapped to the higher-level behavioral categories described in Section~\ref{sec:methods}. This mapping groups semantically related teacher descriptions into interpretable behavioral categories that capture strengths, difficulties, and contextual observations in student narratives. The complete mapping between granular themes and higher-level behavioral categories is provided in Appendix~\ref{sec:appendix2}.

Table~\ref{tab:theme_freq} compares the average number of themes per narrative for selected behavioral categories across control and structured false-negative groups. Several difficulty-related themes appear substantially more often in structured false-negative ADHD cases than in controls. In particular, \textit{behavioral difficulty} and \textit{attention difficulty} are markedly more prevalent among structured false negatives, suggesting that teachers frequently describe behavioral challenges even when standardized rating scales fail to detect strong ADHD signals. Conversely, narratives of control students contain higher frequencies of several strength-related themes, including \textit{academic strength} and \textit{social-emotional strength}. Control narratives also include more references to \textit{family support}, a theme that does not appear in the structured false-negative group.

\begin{table}[t]
\centering
\small
\setlength{\tabcolsep}{4pt}
\begin{tabular}{lccc}
\hline
\textbf{Theme} & \textbf{Control} & \textbf{FN} &\textbf{Diff} \\
\hline
Behavioral difficulty & 0.21 & 0.62 & +0.41 \\
Attention difficulty & 0.24 & 0.50 & +0.26 \\
Academic difficulty & 0.33 & 0.54 & +0.21 \\
Neutral context & 0.09 & 0.29 & +0.20 \\
Family difficulty & 0.03 & 0.00 & -0.03 \\
Behavioral strength & 0.64 & 0.46 & -0.18 \\
Social-emotional difficulty & 0.61 & 0.42 & -0.19 \\
Family support & 0.24 & 0.00 & -0.24 \\
Academic strength & 1.18 & 0.88 & -0.30 \\
Social-emotional strength & 0.85 & 0.46 & -0.39 \\

\hline
\end{tabular}
\caption{Average theme frequency per narrative for selected behavioral categories in control and structured false-negative groups. Positive differences indicate higher prevalence in structured false negatives.}
\label{tab:theme_freq}
\end{table}

To further examine overall narrative patterns, we compare the total number of strength-related and difficulty-related themes appearing in each narrative, excluding neutral contextual descriptions. As shown in Table~\ref{tab:theme_balance}, structured false-negative ADHD cases contain significantly more difficulty-related themes and substantially fewer strength-related themes than control narratives. These findings provide qualitative insight into the behavioral signals captured by teacher narratives. Although structured false-negative ADHD cases exhibit relatively mild scores in standardized rating scales, teachers frequently describe attention-related and behavioral difficulties in free-text narratives for these students. At the same time, control narratives contain substantially more descriptions of academic and social-emotional strengths.

\begin{table}[t]
\centering
\small
\setlength{\tabcolsep}{4pt}
\begin{tabular}{lcccc}
\hline
\textbf{Metric} & \textbf{Ctrl Avg.} & \textbf{FN Avg.} & \textbf{Diff} & $\textbf{p}$ \\
\hline
Strength themes & 2.91 & 1.79 & -1.12 & 0.0009 \\
Difficulty themes & 1.42 & 2.08 & +0.66 & 0.0128 \\
\hline
\end{tabular}
\caption{Comparison of the mean number of strength and difficulty themes per student between control and structured false-negative (FN) groups. Diff denotes the difference in group means (FN $-$ control). $p$-values are computed using two-sided Mann--Whitney tests.}
\label{tab:theme_balance}
\end{table}

Collectively, with the quantitative analyses presented in Section~\ref{exp:5.3}, these results suggest that narrative descriptions capture meaningful behavioral patterns that may not be fully reflected in standardized teacher rating scales. Teacher narratives therefore provide complementary behavioral evidence for identifying ADHD-related difficulties that remain difficult to detect using structured rating items alone.

\section{Conclusion and Future Work}
\label{sec:conclusion}

This study examines how structured behavioral rating scales and teacher narratives capture different signals related to ADHD. While classifiers trained on structured CTRS-R:S ratings achieve strong overall performance, our analysis reveals that a subset of ADHD cases remains difficult to detect using structured scores alone. These structured false-negative cases exhibit rating patterns that overlap substantially with control students, suggesting that standardized behavioral scales may not fully capture all observable classroom difficulties.

In contrast, narrative-based models trained on teacher descriptions show stronger discriminative ability for these cases. Quantitative error analyses demonstrate that narrative signals can distinguish structured false-negative ADHD cases from controls more effectively than structured ratings. Furthermore, LLM-assisted theme analysis reveals systematic differences in how teachers describe these students: narratives of structured false-negative ADHD cases contain more behavioral difficulty descriptions and fewer strength-related descriptions compared to control narratives. Overall, our findings highlight the value of combining structured behavioral assessments with narrative teacher reports when analyzing behavioral signals related to ADHD. More broadly, this study demonstrates how NLP methods can help uncover clinically meaningful behavioral patterns embedded in open-ended teacher narratives.

Future work can extend this study in several directions. Collecting larger and more balanced datasets—particularly by increasing the number of control students—would enable more robust statistical analyses and more reliable evaluation of narrative-based models. In addition, while this study relies on the Llama 3.3 70B model for theme extraction, exploring alternative language models, especially those with stronger Turkish language capabilities, may further improve theme identification. Finally, expanding this analysis to teacher evaluation data from other educational systems or languages would allow future studies to examine whether similar behavioral patterns emerge across different cultural and educational contexts.

\section*{Limitations}

This study has several limitations related to data size, modeling choices, and generalizability. The dataset is relatively small and imbalanced (166 ADHD vs. 33 control students), which limits the complexity of models that can be reliably trained and evaluated. As a result, the narrative classification analysis relies on TF-IDF representations with logistic regression rather than more expressive neural architectures. While this approach provides a stable and interpretable baseline for small datasets, larger corpora would enable the use of models that capture richer linguistic patterns in teacher narratives.

In addition, teacher narratives are inherently subjective and may reflect individual reporting styles, classroom expectations, or teacher perceptions, which can introduce variability in how student behaviors are described. Because the narratives analyzed in this study originate from a single educational and cultural context, the behavioral patterns identified here may not fully generalize to other educational systems, languages, or reporting environments. Finally, although LLM-assisted theme extraction provides a structured way to summarize narrative signals, the mapping of extracted themes to higher-level behavioral categories involves expert-informed design decisions, which may introduce some degree of subjectivity.

\bibliography{custom}

\renewcommand{\thetable}{A\arabic{table}}
\renewcommand{\thefigure}{A\arabic{figure}}
\setcounter{table}{0}  
\setcounter{figure}{0}  

\clearpage 
\appendix

\section{Appendix}
\label{sec:appendix}

\subsection{Original Theme Extraction Prompt}
\label{sec:appendix1}

\begin{figure}[H]
\centering
\begin{tcolorbox}
\footnotesize
\textbf{SYSTEM:} Sen klinik tanı koymazsın. \\
Öğretmen anlatılarından gözlenebilir davranışları kısa TEMALAR olarak çıkarırsın. \\

\textbf{Kurallar:} \\
- Tanı/etiket yazma: DEHB/ADHD, tanı, rapor, ilaç vb. \\
- Her tema 2–6 kelime olmalı. \\
- Her tema açıkça bir GÜÇ (pozitif) veya ZORLUK (negatif) ifade etmeli. \\
- Nötr tema üretme. \\
- Sadece metinde kanıtı olan temaları yaz. \\
- Çıktı yalnızca geçerli JSON olsun. \\

\textbf{Örnek temalar:} \\
\textit{"sosyal uyum güçlü"} \\
\textit{"arkadaş ilişkilerinde zorluk"} \\

\textbf{USER:} 
Aşağıdaki öğretmen anlatısını oku ve HER BÖLÜM için en fazla 2 tema çıkar. \\

Bölümler: \\
1) OKUL BAŞARISI \\
2) KİŞİLİĞİ / RUHSAL \\
3) SOSYAL İLİŞKİLER \\
4) AİLE \\

\textbf{JSON formatı:}

\begin{quote}
\ttfamily
\{\\
\hspace*{1em}"themes": [\\
\hspace*{2em}\{\\
\hspace*{3em}"tema": "...",\\
\hspace*{3em}"valans": "pozitif|negatif",\\
\hspace*{3em}"alan": "okul|kişisel|sosyal|aile"\\
\hspace*{2em}\},\\
\hspace*{2em}...\\
\hspace*{1em}]\\
\}
\end{quote}

\textbf{Ek kurallar:}
Her alan için en fazla 2 tema, toplam en fazla 8 tema, ve temalar metnin içeriğine dayanmalı. \\

\textbf{Metin:} \texttt{<<< \{TEXT\} >>>}
\end{tcolorbox}
\caption{Original Turkish prompt used for LLM-assisted theme extraction from teacher narratives.}
\label{fig:prompt2}
\end{figure}

\subsection{Complete Mapping from Granular Themes to Higher-Level Behavioral Categories}
\label{sec:appendix2}

To improve transparency and reproducibility, we provide the complete mapping between the granular behavioral themes extracted from teacher narratives and the higher-level behavioral categories used in the analysis. The themes are originally written in Turkish and are accompanied by English glosses to facilitate interpretation for non-Turkish-speaking readers. The glosses represent approximate semantic translations rather than literal word-for-word translations.

The higher-level categories are:
\textit{attention difficulty},
\textit{academic difficulty},
\textit{behavioral difficulty},
\textit{social-emotional difficulty},
\textit{academic strength},
\textit{behavioral strength},
\textit{social-emotional strength},
\textit{family support},
\textit{family difficulty}, and
\textit{neutral context}.

For readability, the complete mapping is presented across four tables grouped by behavioral polarity and thematic domain. Tables~\ref{tab:theme_mapping_difficulty_1} and~\ref{tab:theme_mapping_difficulty_2} list difficulty-related themes, while Tables~\ref{tab:theme_mapping_strength_1} and~\ref{tab:theme_mapping_strength_2} present strength-related themes. Themes categorized as \textit{neutral context} are not included in the tables, as they do not express a clear strength or difficulty but instead provide descriptive background information.

\begin{table*}[t]
\centering
\small
\begin{tabular}{p{5.2cm} p{5.8cm} p{4cm}}
\hline
\textbf{Turkish (original) theme} & \textbf{English gloss} & \textbf{Higher-level category} \\
\hline
kafası karışık & confused / mentally scattered & attention difficulty \\
dağınıklık & disorganization & attention difficulty \\
dağınık davranış & disorganized behavior & attention difficulty \\
dikkat dağınıklığı & distractibility & attention difficulty \\
nadiren dalgın & occasionally absent-minded & attention difficulty \\
dikkat eksikliği & lack of attention & attention difficulty \\
dikkatsizlik nedeniyle hatalar & mistakes due to inattention & attention difficulty \\
derste dalmak & zoning out in class & attention difficulty \\
dalgın ve dağınık & absent-minded and disorganized & attention difficulty \\
ders esnasında dikkati dağılabiliyor & attention drifts during class & attention difficulty \\
dikkatli olduğunda başarılı & successful when attentive & attention difficulty \\
ilgi duyduğunda başarılı & successful when interested & attention difficulty \\
konsantrasyon sorunu & concentration problems & attention difficulty \\
dalgın ve dağınıktır & absent-minded and disorganized & attention difficulty \\
sıkılır ve motive olmuyor & gets bored and unmotivated & attention difficulty \\
dikkat ve odaklanma sorunu & attention and focus problems & attention difficulty \\
ilgisini çekme ihtiyacı & needs to be engaged / interested & attention difficulty \\

akademik zorluklar & academic difficulties & academic difficulty \\
yönergeleri takip etmekte zorluk & difficulty following instructions & academic difficulty \\
vasat başarı & average achievement & academic difficulty \\
daha başarılı olabilir & could perform better & academic difficulty \\
resim becerisi yok & lacks drawing skills & academic difficulty \\
kitap okumaya karşı ilgisizlik & lack of interest in reading books & academic difficulty \\
problem çözmede zorlanıyor & struggles with problem solving & academic difficulty \\
yazma yavaş & writes slowly & academic difficulty \\
bazı konularda zorluk & difficulty in some subjects & academic difficulty \\
yönerge uygulamada zorluk & difficulty applying instructions & academic difficulty \\
zaman yönetimi zorluğu & difficulty with time management & academic difficulty \\
özümseme güçlüğü & difficulty with comprehension / internalization & academic difficulty \\
yazı yazma zorluğu & difficulty with writing & academic difficulty \\
ezberleme gücü zayıf & weak memorization ability & academic difficulty \\
sayısal derslerde zayıf & weak in quantitative subjects & academic difficulty \\
okuma performansı değişken & inconsistent reading performance & academic difficulty \\
ödevlere karşı isteksiz & reluctant to do homework & academic difficulty \\
anlamada zorluk & difficulty understanding & academic difficulty \\
matematikte zorluk & difficulty in mathematics & academic difficulty \\
başarı dalgalı & inconsistent performance & academic difficulty \\
yavaş iş yapma & works slowly & academic difficulty \\
işlem hatası & computational error & academic difficulty \\
sayısal derslerde zorluk & difficulty in quantitative subjects & academic difficulty \\
ders başarısında zorluk & difficulty with academic performance & academic difficulty \\
ders başarısı düşük & low academic performance & academic difficulty \\
ders çalışma zorluğu & difficulty studying & academic difficulty \\
\hline
\end{tabular}
\caption{Granular difficulty-related themes and their mapping to higher-level categories (Part I: attention and academic difficulties).}
\label{tab:theme_mapping_difficulty_1}
\end{table*}

\begin{table*}[t]
\centering
\small
\begin{tabular}{p{5.6cm} p{5.6cm} p{4cm}}
\hline
\textbf{Turkish (original) theme} & \textbf{English gloss} & \textbf{Higher-level category} \\
\hline

davranışsal zorluklar & behavioral difficulties & behavioral difficulty \\
kendini sıkar & puts pressure on self / tenses self up & behavioral difficulty \\
öfke kontrolü zorluğu & difficulty controlling anger & behavioral difficulty \\
öfke kontrolü zor & poor anger control & behavioral difficulty \\
fazla müdahaleci & overly intrusive / interferes too much & behavioral difficulty \\
neşeli ve inatçı & cheerful but stubborn & behavioral difficulty \\
fiziksel becerileri zayıf & weak physical skills & behavioral difficulty \\
inatçı davranış & stubborn behavior & behavioral difficulty \\
davranışları değişken & inconsistent behavior & behavioral difficulty \\
oldukça hareketli ve öfkeli & quite hyperactive and angry & behavioral difficulty \\
inatçı ve hırçın & stubborn and irritable & behavioral difficulty \\
uzlaşılmaz davranışlar & uncooperative behavior & behavioral difficulty \\
sinirli davranışlar & irritable behavior & behavioral difficulty \\
alıngan davranışlar & overly sensitive behavior & behavioral difficulty \\
inatçı & stubborn & behavioral difficulty \\
ben merkezci tavırlar & self-centered attitudes & behavioral difficulty \\
daha sakin davranış & relatively calmer behavior & behavioral difficulty \\
fazla konuşma & excessive talking & behavioral difficulty \\
kuralları bozma eğilimi & tendency to break rules & behavioral difficulty \\
inatçı ve yaramaz & stubborn and mischievous & behavioral difficulty \\
yerinden kalkma eğilimi & tendency to leave seat & behavioral difficulty \\
paylaşımda zorluk & difficulty sharing & behavioral difficulty \\

arkadaş edinme zorlukları & difficulty making friends & social-emotional difficulty \\
sosyalleşmede pasif & passive in socialization & social-emotional difficulty \\
utangaç & shy & social-emotional difficulty \\
arkadaş seçmekte zorluk & difficulty choosing friends & social-emotional difficulty \\
zaman zaman arkadaşlarına inatlaşabiliyor & sometimes becomes oppositional with peers & social-emotional difficulty \\
çekingen kişilik & timid personality & social-emotional difficulty \\
içe kapanıklık & social withdrawal & social-emotional difficulty \\
pasif ilişki & passive interpersonal style & social-emotional difficulty \\
sosyalleşme isteksizliği & reluctance to socialize & social-emotional difficulty \\
korktukça sessizleşiyor & becomes quiet when afraid & social-emotional difficulty \\
kaygılı ve korkak & anxious and fearful & social-emotional difficulty \\
sevdiği öğretmenlerle rahat & comfortable only with favored teachers & social-emotional difficulty \\
ilgi bekler & seeks attention / expects attention & social-emotional difficulty \\
kırılgan & emotionally fragile & social-emotional difficulty \\
duygusal ve alıngan & emotional and easily offended & social-emotional difficulty \\
duygusal içe dönüklük & emotional introversion & social-emotional difficulty \\
hassas öğrenci & sensitive student & social-emotional difficulty \\
duygusal ve sessiz & emotional and quiet & social-emotional difficulty \\
duygusal dalgalanma & emotional fluctuation & social-emotional difficulty \\
hassas & sensitive & social-emotional difficulty \\
hassas ve sıkılgan & sensitive and easily bored / withdrawn & social-emotional difficulty \\
hassas ve alıngan & sensitive and easily offended & social-emotional difficulty \\
hassas ve çabuk küsebilen & sensitive and easily upset & social-emotional difficulty \\
eleştirilere karşı duyarlı & sensitive to criticism & social-emotional difficulty \\
duygusal ve içe dönük & emotional and introverted & social-emotional difficulty \\
duygusal hassasiyet & emotional sensitivity & social-emotional difficulty \\
hassas ve sevgi dolu & sensitive and affectionate & social-emotional difficulty \\
sessiz bir mizacı & quiet temperament & social-emotional difficulty \\
hata yapma korkusu var & fear of making mistakes & social-emotional difficulty \\

endişeli aile & anxious family & family difficulty \\

\hline
\end{tabular}
\caption{Granular difficulty-related themes and their mapping to higher-level categories (Part II: behavioral, social-emotional, and family difficulties).}
\label{tab:theme_mapping_difficulty_2}
\end{table*}

\begin{table*}[t]
\centering
\small
\begin{tabular}{p{5cm} p{6.2cm} p{4cm}}
\hline
\textbf{Turkish (original) theme} & \textbf{English gloss} & \textbf{Higher-level category} \\
\hline

akademik başarı & academic success & academic strength \\
başarılı & successful & academic strength \\
başarılı öğrenci & successful student & academic strength \\
sanatsal yetenek & artistic talent & academic strength \\
tüm derslerde başarılı & successful across all subjects & academic strength \\
başarı artışı & improvement in performance & academic strength \\
ilgi gösterir & shows interest & academic strength \\
eksiklerini tamamlama çaba & effort to make up deficiencies & academic strength \\
okul başarısı yüksek & high school performance & academic strength \\
derslerinde başarılı & successful in coursework & academic strength \\
yönergelere uyar & follows instructions & academic strength \\
genel olarak başarılı & generally successful & academic strength \\
okuma becerisi iyi & good reading skills & academic strength \\
okuma becerisi güçlü & strong reading skills & academic strength \\
sanat yeteneği var & has artistic ability & academic strength \\
problem çözme yeteneği gelişmiş & well-developed problem-solving ability & academic strength \\
ders başarısı yüksek & high academic performance & academic strength \\
el becerileri iyi & good fine motor / hand skills & academic strength \\
dersi derste kavrayabilme & able to grasp lessons during class & academic strength \\
araştırmacı ve başarılı & inquisitive and successful & academic strength \\
matematik başarısı yüksek & high math achievement & academic strength \\
sözel becerisi güçlü & strong verbal ability & academic strength \\
futbolda başarılı & successful in football & academic strength \\
sözel ve sayısalda başarılı & successful in both verbal and quantitative subjects & academic strength \\
akademik başarısı yüksek & high academic achievement & academic strength \\
sanat ve spor yeteneği var & has talent in art and sports & academic strength \\
okuduğunu anlama becerisi güzel & good reading comprehension & academic strength \\
derslerde başarılı & successful in classes & academic strength \\
anlama becerisi iyi & good comprehension ability & academic strength \\
gayretli öğrenci & hardworking student & academic strength \\
spora yeteneği var & has athletic ability & academic strength \\
okuma yazma becerileri güçlü & strong literacy skills & academic strength \\
etkinliklere katılımı başarılı & successful participation in activities & academic strength \\
başarı düzeyi iyidir & achievement level is good & academic strength \\
derslere katılır & participates in class & academic strength \\
matematik başarısı & math achievement & academic strength \\
başarı düzeyi yüksek & high achievement level & academic strength \\
sözel yeteneği yüksek & high verbal ability & academic strength \\
sanatsal etkinliklere ilgi & interest in artistic activities & academic strength \\
matematik dersinde başarılı & successful in math class & academic strength \\
sorgulama becerisi güçlü & strong questioning / inquiry skills & academic strength \\
sözel derslerde başarılı & successful in verbal subjects & academic strength \\
meraklı ve sorgulayıcı & curious and inquisitive & academic strength \\
öğrenme isteği var & shows willingness to learn & academic strength \\
matematik yeteneği güçlü & strong mathematical ability & academic strength \\

aile ilgili & family is involved & family support \\
aile desteği & family support & family support \\
aile içi uyum & harmony within the family & family support \\
aile uyumlu & supportive family environment & family support \\
aile içi ilişkileri iyi & good family relationships & family support \\
aile ilişkileri iyi & positive family relations & family support \\

\hline
\end{tabular}
\caption{Granular strength-related themes and their mapping to higher-level categories (Part I: academic strength and family support).}
\label{tab:theme_mapping_strength_1}
\end{table*}

\begin{table*}[t]
\centering
\small
\begin{tabular}{p{5.2cm} p{5.2cm} p{4.2cm}}
\hline
\textbf{Turkish (original) theme} & \textbf{English gloss} & \textbf{Higher-level category} \\
\hline

titiz ve tertipli & neat and orderly & behavioral strength \\
titizlik & neatness / carefulness & behavioral strength \\
söz dinler & listens to instructions / is obedient & behavioral strength \\
yaşından büyük davranıyor & behaves older than their age & behavioral strength \\
güler yüzlü & cheerful / smiling & behavioral strength \\
neşeli ve hareketli & cheerful and energetic & behavioral strength \\
gayretli ve dürüst & hardworking and honest & behavioral strength \\
saygılı öğrenci & respectful student & behavioral strength \\
büyüklere karşı saygılı & respectful toward adults & behavioral strength \\
yaşından büyük davranış & mature behavior beyond age & behavioral strength \\
sorumluluk sahibi & responsible & behavioral strength \\
yaşına göre olgun & mature for their age & behavioral strength \\
kendini kolay ifade eden & expresses self easily & behavioral strength \\
ağırbaşlı ve olgun & serious and mature & behavioral strength \\
çok düzenlidir & very organized & behavioral strength \\
kurallara uyan & follows rules & behavioral strength \\
saygılı davranış & respectful behavior & behavioral strength \\
dürüsttür & honest & behavioral strength \\
söz dinleme & compliance with instructions & behavioral strength \\
işini iyi yapmak ister & wants to do things well & behavioral strength \\
kurallara sadık & adheres to rules & behavioral strength \\
konuşkan ve olgun & talkative and mature & behavioral strength \\
kurallara uyma & rule-following & behavioral strength \\
hayal gücü yüksek & strong imagination & behavioral strength \\

sevilen öğrenci & well-liked student & social-emotional strength \\
arkadaşları tarafından sevilen & liked by classmates & social-emotional strength \\
lider ruhlu & shows leadership qualities & social-emotional strength \\
arkadaşları tarafından sevilir & liked by peers & social-emotional strength \\
arkadaşları ile arası iyi & gets along well with peers & social-emotional strength \\
neşeli & cheerful & social-emotional strength \\
neşeli arkadaş & cheerful friend & social-emotional strength \\
neşeli ve güler yüzlü & cheerful and smiling & social-emotional strength \\
sosyal uyum & good social adjustment & social-emotional strength \\
duygusal dengesi iyi & emotionally balanced & social-emotional strength \\
arkadaş çevresinde popüler & popular among peers & social-emotional strength \\
sevilir & well liked & social-emotional strength \\
liderlik özellikleri var & has leadership qualities & social-emotional strength \\
arkadaşları ile iyi geçiniyor & gets along well with friends & social-emotional strength \\
neşeli ve heyecanlı kişilik & cheerful and enthusiastic personality & social-emotional strength \\
liderlik özellikleri & leadership traits & social-emotional strength \\
ilgi ve paylaşım gösteriyor & shows care and sharing behavior & social-emotional strength \\
neşeli ve esprili & cheerful and humorous & social-emotional strength \\
arkadaş canlısı & friendly & social-emotional strength \\
sakin ve sevilen & calm and well-liked & social-emotional strength \\
iletişime açık & open to communication & social-emotional strength \\
liderlik özelliği & leadership trait & social-emotional strength \\
liderlik özelliği var & has leadership ability & social-emotional strength \\
neşeli kişilik & cheerful personality & social-emotional strength \\
iletişimi güçlü & strong communication skills & social-emotional strength \\
atılgan & assertive & social-emotional strength \\
atılgan ve neşeli & assertive and cheerful & social-emotional strength \\
sosyal & social & social-emotional strength \\
öğretmen ile iyi iletişim & good communication with the teacher & social-emotional strength \\
duygu kontrolü gelişimi & developed emotional regulation & social-emotional strength \\
bağ kurma yeteneği & ability to form relationships & social-emotional strength \\
sosyal bir öğrenci & socially engaged student & social-emotional strength \\
neşeli ve sevilen & cheerful and well-liked & social-emotional strength \\

\hline
\end{tabular}
\caption{Granular strength-related themes and their mapping to higher-level categories (Part II: behavioral and social-emotional strengths).}
\label{tab:theme_mapping_strength_2}
\end{table*}

\end{document}